\documentclass[11pt]{article}
\usepackage{blindtext}
\usepackage[preprint,abbrvbib]{jmlr2e}
\usepackage{enumitem}
\usepackage{amsmath}

\usepackage[T1]{fontenc}
\usepackage{listings}
\usepackage{xcolor}
\usepackage{courier}
 
\definecolor{codegreen}{rgb}{0.15,0.6,0.3}
\definecolor{codegray}{rgb}{0.5,0.5,0.5}
\definecolor{codepurple}{rgb}{0.73,0.13,0.13}
\definecolor{backcolour}{rgb}{1,1,1}
 
\lstdefinestyle{mystyle}{
    backgroundcolor=\color{backcolour},   
    commentstyle=\itshape\color{codegreen},
    keywordstyle=\bfseries\color{magenta},
    numberstyle=\tiny\color{codegray},
    stringstyle=\color{codepurple},
    basicstyle=\footnotesize\ttfamily,
    breakatwhitespace=false,
    breaklines=true,
    captionpos=b,
    keepspaces=true,
    numbers=left,
    numbersep=5pt,
    showspaces=false,
    showstringspaces=false,
    showtabs=false,                  
    tabsize=1
}

\newcommand{\name}{\texttt{ZhiJian}\xspace}
\DeclareMathOperator{\st}{s.t.}

\ShortHeadings{ZhiJian: A Unifying and Rapidly Deployable Toolbox for Pre-trained Model Reuse}{Zhang, Ren, Yi, Wang, Zhan, and Ye}
\firstpageno{1}

\begin{document}
\lstset{language=Python}

\title{ZhiJian: A Unifying and Rapidly Deployable Toolbox\\for Pre-trained Model Reuse}

\author{ \\
\name Yi-Kai Zhang \email zhangyk@lamda.nju.edu.cn \\
\name Lu Ren \email renl@lamda.nju.edu.cn \\
\name Chao Yi \email yic@lamda.nju.edu.cn \\
\name Qi-Wei Wang \email wangqiwei@lamda.nju.edu.cn \\
\name De-Chuan Zhan \email zhandc@lamda.nju.edu.cn \\
\name Han-Jia Ye \email yehj@lamda.nju.edu.cn \\
\addr State Key Laboratory for Novel Software Technology, Nanjing University, Nanjing, 210023, China}
\editor{}

\maketitle

\begin{abstract}
The rapid expansion of foundation pre-trained models and their fine-tuned counterparts has significantly contributed to the advancement of machine learning. Leveraging pre-trained models to extract knowledge and expedite learning in real-world tasks, known as ``Model Reuse'', has become crucial in various applications. Previous research focuses on reusing models within a certain aspect, including reusing model weights, structures, and hypothesis spaces.
This paper introduces {\name}, a comprehensive and user-friendly toolbox for model reuse, utilizing the PyTorch backend. {\name} presents a novel paradigm that unifies diverse perspectives on model reuse, encompassing target architecture construction with PTM, tuning target model with PTM, and PTM-based inference. This empowers deep learning practitioners to explore downstream tasks and identify the complementary advantages among different methods.
{\name} is readily accessible at \url{https://github.com/zhangyikaii/lamda-zhijian}, facilitating seamless utilization of pre-trained models and streamlining the model reuse process for researchers and developers.
\end{abstract}


\section{Introduction}
The rapid progress in deep learning techniques~\citep{he2016deep,DevlinCLT19,RadfordKHRGASAM21,DBLP:conf/eccv/JiaTCCBHL22} has led to the emergence of numerous open-source Pre-Trained Models (PTMs) provided by platforms like PyTorch~\citep{benoit_pytorch:_2019}, TensorFlow~\citep{abadi_tensorflow:_2016}, and HuggingFace Transformers~\citep{wolf_transformers:_2020}. Leveraging these PTMs for specific tasks empowers them to handle targeted objectives effectively, resulting in the creation of valuable model resources that contribute to the growth of the machine-learning community. As a result, the practice of {\em reusing} PTMs has become increasingly vital in enhancing the capabilities and efficiency of target models.

``Model Reuse'' can be effectively implemented from various aspects, such as adapting the architecture of the target model, customizing the learning process on the target data, or devising an optimized inference strategy to leverage the knowledge of the PTM. Recent approaches, like parameter-efficient fine-tuning~\citep{DBLP:conf/icml/HoulsbyGJMLGAG19,DBLP:conf/acl/GuoRK20,DBLP:conf/iclr/HuSWALWWC22,DBLP:conf/eccv/JiaTCCBHL22,DBLP:journals/corr/abs-2207-07039,DBLP:conf/nips/LianZFW22,DBLP:conf/aaai/JieD23}, hypothesis transfer~\citep{KuzborskijOC13,KuzborskijO17,LiangHF20,YeZJZ21}, distilling the pre-trained knowledge~\citep{ZhouJC03,ZhouJ04,HintonVD15,DBLP:journals/corr/RomeroBKCGB14,DBLP:conf/cvpr/YimJBK17,DBLP:journals/corr/HuangW17a,DBLP:conf/cvpr/ParkKLC19,DBLP:conf/iccv/TungM19,DBLP:conf/iclr/TianKI20,DBLP:journals/pami/YeLZ23}, and merging pre-trained models~\citep{DingZ18,WuLZ19,DBLP:conf/icml/WortsmanIGRLMNF22,DBLP:conf/iclr/AinsworthHS23} have predominantly focused on specific aspects, making it challenging to identify the optimal reuse method for a specific target task.
Furthermore, some current popular libraries for previous model reuse methods are not fully compatible~\citep{pfeiffer2020AdapterHub,peft,llama-efficient-tuning}. They differ significantly in their preprocessing configurations and interfaces. As a result, directly combining these libraries may lead to biased evaluations of methods and hampers deep learning practitioners from seamlessly switching or integrating diverse approaches.

To facilitate a holistic consideration of various model reuse strategies, we categorize model reuse methods into three sequential modules: ``{\em Architect}'', ``{\em Tuner}'', and ``{\em Merger}'', aligning with the stages of model preparation, model learning, and model inference on the target task, respectively.
The ``Architect'' module involves modifying the PTM to fit the target task, and reusing certain parts of the PTM while introducing new learnable parameters with specialized structures.
The ``Tuner'' module trains the target model with guidance from PTM knowledge to expedite the optimization process, \textit{e.g.}, via adjusting objectives, optimizers, or regularizers. Finally, the ``Merger'' module influences the inference phase by either reusing pre-trained features or incorporating adapted logits from the PTM.

Based on the three modules, we present a comprehensive and readily deployable model reuse toolbox called {\name}, built on PyTorch. {\name} streamlines the model reuse process, enables effortlessly exploring various approaches to achieve enhanced performance in specific target tasks, and facilitates the creation of novel reuse methods tailored to the unique characteristics of each task.
{\name} possesses the following key features:

\begin{figure}[t]
    \centering
    \includegraphics[width=1.0\linewidth]{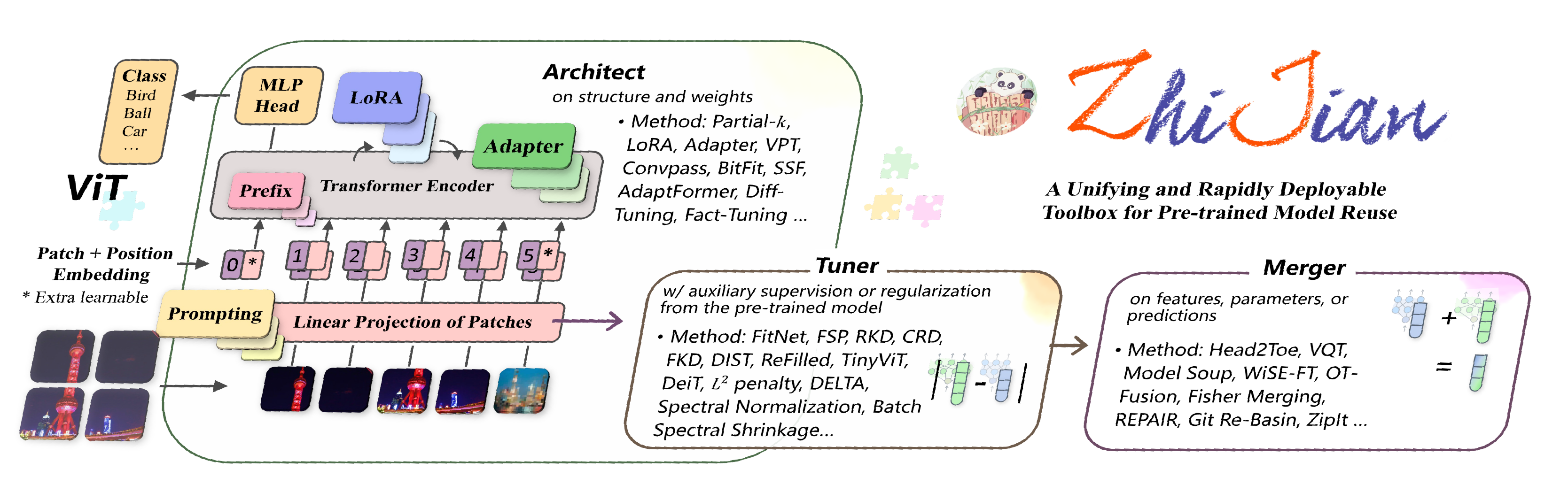}
    \caption{\textbf{The overall structure of {\name}} incorporates three modules, facilitating the seamless workflow conducted by ``architect'', ``tuner'', and ``merger''. 
The architect adapts the architecture and identifies the set of pre-trained weights to be further tuned.
Next, the tuner trains the target model with auxiliary supervision or regularization.
Finally, the merger integrates the features, weights, or predictions of the PTM or variants of the tuned target models.
These processing stages form a cohesive structure, allowing users to easily match and explore combinations.}
    \label{fig:overview}
\end{figure}

\begin{itemize}
\item \textbf{Comprehensive coverage of model reuse stages:} {\name} supports over 30 diverse approaches across the entire model reuse pipeline, which provides a versatile and robust solution for effectively leveraging PTM knowledge. The toolbox includes interfaces to popular PTM libraries from Google, HuggingFace, OpenAI, and ensures compatibility with large foundational visual or language models.

\item \textbf{High flexibility with user-friendly customization options:} Users can effortlessly incorporate additional functionalities, such as adding a \texttt{LoRA} module to a vision transformer, using {\em just one line of configuration code}. The toolbox's components are designed to seamlessly integrate for multi-modal training and multi-model reuse, further enhancing its adaptability to diverse scenarios.

\item \textbf{Easy creation of new reuse strategies:} {\name} features a unifying training interface that enables seamless switching between approaches from various fields. It also establishes standardized evaluation protocols and metrics for assessing performance. Users can readily compare, combine, and explore new techniques on the same benchmark without requiring an in-depth understanding of those specific fields.

\end{itemize}

\section{ZhiJian for Model Reuse}
\label{Pipelines}
We formally define the model reuse task and then provide an overview of the model reuse methods supported by {\name}. The overall structure is summarized in~\autoref{fig:overview}.

\subsection{Preliminary}
The primary objective of the model reuse problem is to leverage the knowledge inherent in the architecture as well as the weights of one or more PTMs {\em without accessing the original upstream pre-training data}.\footnote{In some cases, we have auxiliary pre-training data. While in model reuse, we emphasize the way to utilize the knowledge contained within the {\em PTMs}.} With the help of the PTMs, the learning process of downstream tasks can be significantly expedited and enhanced.

We assume the downstream target task $\mathcal{T}=\left\{\left(\mathbf{x}_i, y_i\right)\right\}_{i=1}^N$ is a classification task with $N$ labeled examples, and our analysis could be extended to other scenarios. The PTM is denoted as $h_{{\boldsymbol{w}}}$, where $h$ and ${\boldsymbol{w}}$ represent the model architecture and the pre-trained weights, respectively. Our goal is to reuse the PTMs to determine the architecture $h'$ as well as learn the weights $\hat{\boldsymbol{w}}$ for the target task.
\begin{align}
    h'_{\hat{\boldsymbol{w}}} \;=\; &\mathcal{M} \left(\left\{ \underset{h_{{\boldsymbol{w}}_j}}{\arg \min } \;\mathbb{E}_{(\mathbf{x},y)} \left[ \ell \left( h'_{{\boldsymbol{w}}_j} \left( \mathbf{x} \right), y \mid \boldsymbol{w} \right) + \varOmega \left( \boldsymbol{w}_j, \boldsymbol{w} \mid \mathbf{x} \right) \right] \right\}_{j=1}^{M}\right),\label{eq:objective}\\
    &\st\; h'_{\boldsymbol{w}_j} = \alpha_j\left( h_{\boldsymbol{w}} \right)\;.\label{eq:architect}
\end{align}

The objective encompasses three primary components. 
$\alpha$ in \autoref{eq:architect} represents the architecture modifier applied to the PTM, serving two key purposes. First, it defines the enhanced model architecture for the target task. This can involve introducing additional modules into the intermediate layers of the PTM's architecture. Furthermore, $\alpha$ identifies the specific weights that should remain fixed during the target task's learning process. This selective freezing of weights ensures that only a subset of weights is fine-tuned, which helps prevent overfitting.
As the modified architecture-dependent downstream weights are constructed using $M$ PTMs, the $j^{\text{th}}$ of them is denoted as $h_{w_j} = \alpha_j(h_{\boldsymbol{w}})$. Notably, multiple configurations can be employed to modify the model's architecture. For instance, tuning the top-$k$ layers or employing diverse initialization strategies. These variations offer flexibility in tailoring the architecture modification to the specific requirements of the target task.

The optimization process in \autoref{eq:objective} involves minimizing a combined objective. The loss function $\ell$ measures the discrepancy between the model's prediction $h'_{\boldsymbol{w}_j}$ and the provided supervision $y$. An additional term $\varOmega$ regularizes the current weights $\boldsymbol{w}_j$ remain close to the pre-trained weights $\boldsymbol{w}$. $\varOmega$ can be implemented either based on the difference in parameters or predictions between models with two sets of parameters.

Before entering the inference phase, the optimized multiple $h'_{\hat{{\boldsymbol{w}}_j}}$ can be further integrated using the operator $\mathcal{M}$ --- through their extracted features, weights, or predictions. If required, the target data can facilitate additional calibration among multiple models.

To sum up, we name the three main components, namely, $\alpha_j \left( \cdot \right) \;\rightarrow\; \ell \left( \cdot \right), \varOmega \left( \cdot \right) \;\rightarrow\; \mathcal{M} \left( \cdot \right)$ in the aforementioned objective as ``Architect'', ``Tuner'', and ``Merger'', respectively, which streamline the model reuse process.

\subsection{Supported Methods}
Regarding the objective described above, {\name} comprehensively implements model reuse methods for ``Architect'', ``Tuner'', and ``Merger''. Specifically, we categorize the supported methods in {\name} as follows.\footnote{We assume the pre-trained model has a Vision Transformer (ViT) backbone~\citep{DosovitskiyB0WZ21}, and some of the methods could also be applied to the ConvNet backbones~\citep{he2016deep} with {\name}.}

\noindent\textbf{Architect}. The architect $\alpha_j$ adapts the architecture and identifies the set of pre-trained weights to be further tuned, including 
\begin{itemize}[itemsep=4pt,topsep=4pt]
    \item \textit{Linear Probing}~\citep{DBLP:conf/nips/YosinskiCBL14}. Learn a linear classifier based on the feature extracted by the PTM; 
    \item \textit{Partial-$k$}~\citep{DBLP:conf/nips/YosinskiCBL14}. Fine-tune the whole model or only the top-$k$ layer; 
    \item \textit{Adapter}~\citep{DBLP:conf/icml/HoulsbyGJMLGAG19}. Insert a two-layer learnable adaptation module between layers or inside blocks in the PTM and apply it to target tasks; 
    \item \textit{LoRA}~\citep{DBLP:conf/iclr/HuSWALWWC22}. Construct a variant of the Adapter with low-rank constraint; 
    \item \textit{Visual Prompt Tuning or Prefix}~\citep{DBLP:conf/eccv/JiaTCCBHL22,DBLP:conf/acl/LiL20}. Concatenate and stack prompts to the inputs between blocks; 
    \item \textit{BitFit}~\citep{DBLP:conf/acl/ZakenGR22}. Modify the bias terms of the query and middle-of-MLP bias terms weights of the PTM; 
    \item \textit{Scaling \& Shifting}~\citep{DBLP:conf/nips/LianZFW22}. Introduce linear transformations and merge the adapted weights via re-parameterization; 
    \item \textit{Diff Pruning}~\citep{DBLP:conf/acl/GuoRK20}. Learn a task-specific ``diff'' vector with a differentiable approximation to the $L_0$-norm penalty to encourage sparsity; 
    \item \textit{Fact-Tuning}~\citep{DBLP:conf/aaai/JieD23}. Tensorize the weights into a single 3D tensor and decompose them into lightweight factors.
\end{itemize}

\noindent\textbf{Tuner}. The tuner $\ell \left( \cdot \right)$ and $\varOmega \left( \cdot \right)$ trains the target model with auxiliary supervision or regularization. There are two main types of tuner. First, the predictions between the current model and the PTM could be matched, such as 
\begin{itemize}[itemsep=4pt,topsep=4pt]
    \item \textit{Knowledge Transfer}~\citep{HintonVD15}. Match the predictions between the target one and the PTM with KL divergence. When two models do not share their class sets, a nearest class mean classifier~\citep{DBLP:journals/pami/MensinkVPC13} could be constructed based on the PTM's feature to approximate the auxiliary supervision~\citep{DBLP:journals/pami/YeLZ23}; 
    \item \textit{FitNet}~\citep{DBLP:journals/corr/RomeroBKCGB14}. Match the hidden layer results across models; 
    \item \textit{FSP}~\citep{DBLP:conf/cvpr/YimJBK17}. Generate the flow of the solution procedure matrix with layer features to encode the pre-trained knowledge and optimize it on the target tasks; 
    \item \textit{RKD}~\citep{DBLP:conf/cvpr/ParkKLC19}. Penalize structural differences in pre-trained and target relations by distance-wise and angle-wise constraint; 
    \item \textit{CRD}~\citep{DBLP:conf/iclr/TianKI20}. Introduce contrastive learning to push the representation closer or apart from the PTM's; 
    \item \textit{ReFilled}~\citep{DBLP:journals/pami/YeLZ23}. Match the comparison relationship between the target model and the PTM. 
\end{itemize}
Another thread directly regularizes the parameters based on the weights of the PTM:
\begin{itemize}[itemsep=4pt,topsep=4pt]
    \item \textit{$L^2$ penalty}. Penalize the $L^2$-norm of the weights when fine-tuning the PTM, which equals a kind of weight decay; 
    \item \textit{$L^2$-SP}~\citep{DBLP:conf/icml/LiGD18}. Penalize the $L^2$-norm of the difference between the target weights and the PTM's weights; 
    \item \textit{Spectral Normalization}~\citep{DBLP:conf/iclr/MiyatoKKY18}. Constrain the spectral norm of each layer to control the Lipschitz constant of the pre-trained knowledge; 
    \item \textit{DELTA}~\citep{DBLP:conf/iclr/LiXWRLH19}. Utilize unactivated channel re-usage to select discriminative features with attention and align behaviors between the target model and the PTM;
    \item \textit{Batch Spectral Shrinkage}~\citep{DBLP:conf/nips/ChenWFLW19}. Apply SVD to the feature matrix and penalize the smallest $k$ singular values to suppress the untransferable components.
\end{itemize}

\noindent\textbf{Merger}. The merger $\mathcal{M}$ integrates the features, weights, or predictions of the PTM or variants of the tuned target models.
The feature merger takes into account the intermediate layer's outputs in addition to the final layer's, which includes
\begin{itemize}[itemsep=4pt,topsep=4pt]
    \item \textit{NCM}~\citep{DBLP:journals/pami/MensinkVPC13}. The nearest class mean classifier could be applied over the features extracted by the PTM~\citep{DBLP:journals/corr/abs-1911-04623};
    \item \textit{Head2Toe}~\citep{DBLP:conf/icml/EvciDLM22}. Select features from intermediate layers of the PTM and train an additional head module for the target task;
    \item \textit{Visual Query Tuning}~\citep{Tu2023Visual}. Learn to combine intermediate representations from the PTM.
\end{itemize}
The prediction merger includes: 
  \begin{itemize}[itemsep=4pt,topsep=4pt]
    \item \textit{Logits Ensemble}. Integrate the output logits of multiple models;
    \item \textit{Probability Ensemble}. Integrate the posterior probability of various models;
    \item \textit{Prediction Ensemble}. Integrate the prediction ({\it e.g.}, the predicted classes) of various PTMs;
  \end{itemize}
The weights merger fuses the weights of the PTM and one or more learned models, which include: 
  \begin{itemize}[itemsep=4pt,topsep=4pt]
      \item \textit{Model Soup}~\citep{DBLP:conf/icml/WortsmanIGRLMNF22}. Average weights of multiple fine-tuned models with various strategies (uniform, greedy, and learned soup); 
      \item \textit{WiSE-FT}~\citep{DBLP:conf/cvpr/WortsmanIKLKRLH22}. Interpolate the weights of the PTM and the fine-tuned models; 
      \item \textit{OT-Fusion}~\citep{DBLP:conf/nips/SinghJ20}. Utilize optimal transport to map neurons from the pre-trained view; 
      \item \textit{Fisher Merging}~\citep{DBLP:conf/nips/MatenaR22}. Leverage the Laplace approximation with the diagonal of each model’s Fisher information to combine the acquired knowledge; 
      \item \textit{REPAIR}~\citep{DBLP:conf/iclr/JordanSSEN23}. Rescale preactivation to reduce linear interpolation barrier to mitigate helpful pre-trained phenomenon; 
      \item \textit{Git Re-Basin}~\citep{DBLP:conf/iclr/AinsworthHS23}. Utilize implicit Sinkhorn differentiation to align weights across the PTMs before averaging; 
      \item \textit{ZipIt}~\citep{DBLP:journals/corr/abs-2305-03053}. Match redundant features and merge within each PTM.
  \end{itemize}

\section{The Workflow, API, and Highlights}
In this section, we summarize the main workflow when using {\name}. 

\begin{figure}[t]
    \centering
    \includegraphics[width=0.9\linewidth]{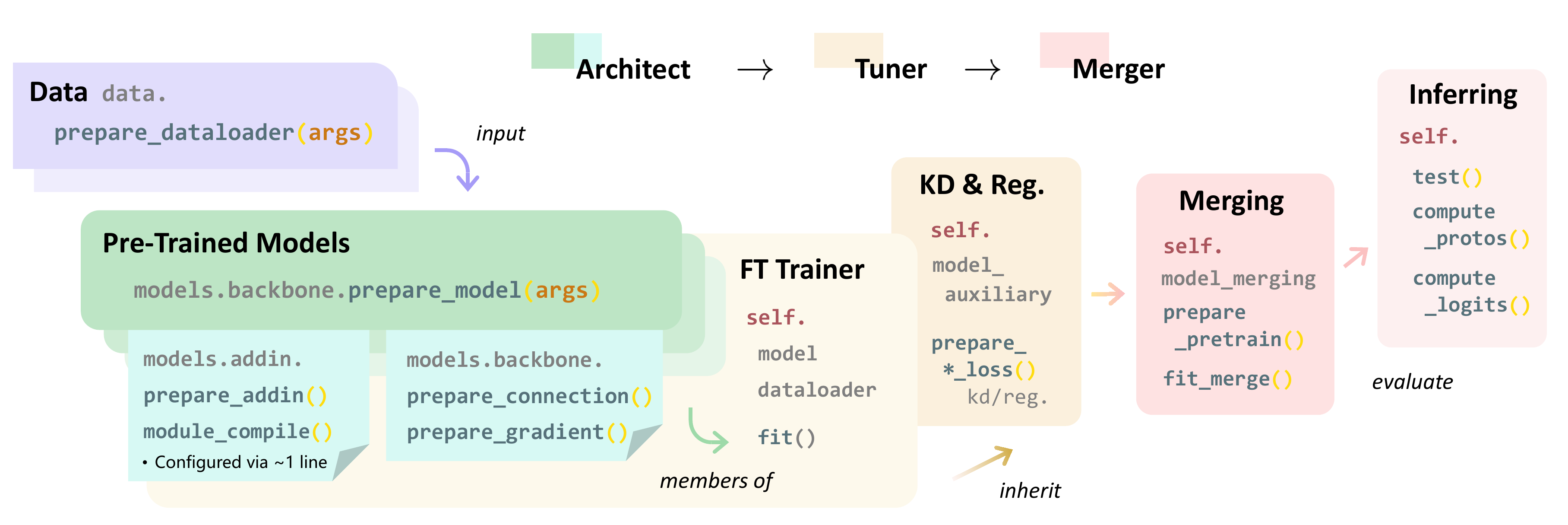}
    \caption{\textbf{The deployment pipelines of {\name}} involve the seamless integration of PTM with advanced structures and downstream data. Diverse tuning trainers \textit{inherit} from the primary one and include the model with data \textit{members}, enabling a convenient integration of concatenated structures and training methodologies.}
    \label{fig:architecture}
\end{figure}

\subsection{The Workflow of Zhijian}
As indicated in~\autoref{eq:objective}, the model takes the target data as input, tuning the (modified) PTMs with an auxiliary objective. Finally, predictions generated by multiple learned models, including the PTM itself, can be merged. Therefore, we have structured {\name} around three steps.

The \verb|data| module establishes the interface for the target data. The core \verb|Trainer| module has a member that represents the PTM, aligning with the architect component. Notably, {\name} provides the interfaces to PTM libraries from Google, HuggingFace, OpenAI, which ensures compatibility with substantial foundational visual or language models.

The Tuner inherits from the \verb|Trainer| class, providing access to an array of auxiliary losses. Finally, the \verb|Merging| module combines various fields of the trainer. Furthermore, an inference module facilitates predictions on test data.

Thanks to the diverse fields available within {\name}, which inherit from the base trainer, the toolbox accommodates shared models and data interfaces while maintaining the modularity of decoupled components. This structural design empowers {\name} to seamlessly integrate diverse components as building blocks.
The whole workflow is illustrated in~\autoref{fig:architecture}.

\subsection{A Usage Example of Zhijian}

The API design of {\name} is intentionally crafted in a manner reminiscent of scikit-learn~\citep{scikit-learn}, fostering a user-friendly experience for deep learning practitioners, as exemplified in Code Listing~\ref{listing:api_example}.

In the initial step, the configurations of the ``architect-tuner-merger'' modules are provided in \verb|args| with various parameters. For instance, the modification of the target model's structure based on the PTM can be achieved through a concise one-liner using the parameter \verb|config_in_1_min_blitz| (which can be further configured in an advanced manner with \verb|.yaml| files). Moreover, the \verb|training_mode| parameter corresponds to the tuner component, while the \verb|merging_func| parameter facilitates the selection of distinct model merging methods.
The advantages of those methods in ``architect'', ``tuner'', or ``merger'' could be incorporated based on the previous configurations. 

Subsequently, the \verb|prepare_trainer| method is employed, where \verb|model| and \verb|train_loader| parameters initiate the reuse of the PTM on the target dataset. The actual training and testing procedures are seamlessly executed by invoking the \verb|fit| and \verb|test| functions.

Moreover, users can delve into advanced functionalities, such as crafting custom auxiliary modules akin to LoRA~\citep{DBLP:conf/iclr/HuSWALWWC22}. This process merely involves implementing relevant data input and output functions, which can be effortlessly incorporated into the existing framework via {\name}'s configurations, without necessitating an in-depth grasp of the underlying mechanics.

\newpage
\begin{lstlisting}[caption={A toy example of the \texttt{ZhiJian} training API. The selection of various modules is achieved by modifying the arguments. As demonstrated in the code, we pass \texttt{args} such as \texttt{config\_in\_1\_min\_blitz}, \texttt{training\_mode}, and \texttt{merging\_func} to configure our task. A unified \texttt{trainer} class is prepared to invoke diverse methods. Accordingly, the \texttt{fit} and \texttt{test} interfaces are conducted for the training and inference processes.}, label={listing:api_example}]
from zhijian.trainers import get_args, prepare_trainer
args = get_args(
  # Architect: configuration with one-line setup, also support .yaml structure
  config_in_1_min_blitz='(LoRA.adapt):->(blocks[0:12].attn.qkv){inout1}->...',
  # Tuner: training mode, or 'knowledge_distillation', 'regularization' ...
  training_mode='finetune',
  # Merger: inference phase, or set to 'base', 'ncm', 'ot_merging' ...
  merging_func='soup',
  # File paths of pre-trained weights, support multiple models
  pretrained_weights=['./m1.pt', './m2.pt', ...]
  ...
) # Flexibly combine different args config to try out new methods
trainer = prepare_trainer(
  args,
  model=PTM,
  train_loader=dataset_loader,
  ...
) # Contruct the downstream trainer
trainer.fit()
trainer.test()
\end{lstlisting}

The modular structure of {\name} empowers users to readily tweak configurations to amalgamate reuse methods spanning different domains. This facilitates practitioners in promptly experimenting with innovative concepts across various target tasks. By virtue of this flexibility, users can readily compare, combine, and explore novel techniques within a standardized benchmark framework, even in fields they may not be deeply familiar with.

\section{Conclusion and Future Development}

We have introduced {\name}, a comprehensive and user-friendly toolbox designed for pre-trained model reuse. With its unified interface, extensive coverage of Pre-Trained Models (PTMs)/reuse methods, and standardized evaluation protocols, {\name} provides a platform for researchers to explore, compare, and apply model reuse techniques effectively. {\name} may serve as a valuable resource for the deep learning community, contributing to the advancement of model reuse research and its applications.

In the future, {\name} can expand its offerings, improve documentation, and actively engage with the community to enhance the toolbox's usability and address the evolving needs of the research and industry communities.



\vskip 0.2in
\bibliography{sample}

\end{document}